\documentclass[conference]{IEEEtran}
\IEEEoverridecommandlockouts
\usepackage{cite}
\usepackage{amsmath,amssymb,amsfonts}
\usepackage{algorithmic}
\usepackage{graphicx}
\usepackage{textcomp}
\usepackage{xcolor}
\def\BibTeX{{\rm B\kern-.05em{\sc i\kern-.025em b}\kern-.08em
    T\kern-.1667em\lower.7ex\hbox{E}\kern-.125emX}}

\usepackage{url}
\usepackage{soul}

\begin{document}


\title{Clustering as an Evaluation Protocol for Knowledge Embedding Representation of Categorised Multi-relational Data in the Clinical Domain
}

\author{
\IEEEauthorblockN{Jianyu Liu}
\IEEEauthorblockA{\textit{Institute of Health Informatics} \\
\textit{University College London}\\
London, UK \\
jianyu.liu@ucl.ac.uk}
\and
\IEEEauthorblockN{Hegler Tissot}
\IEEEauthorblockA{\textit{Institute of Health Informatics} \\
\textit{University College London}\\
London, UK \\
h.tissot@ucl.ac.uk}

}

\maketitle

\begin{abstract}
Learning knowledge representation is an increasingly important technology applicable in many domain-specific machine learning problems. 
We discuss the effectiveness of traditional \emph{Link Prediction} or \emph{Knowledge Graph Completion} evaluation protocol when embedding knowledge representation for categorised multi-relational data in the clinical domain. 
Link prediction uses to split the data into training and evaluation subsets, leading to loss of information along training and harming the knowledge representation model accuracy. 
We propose \emph{Clustering Evaluation Protocol} as a replacement alternative to the traditionally used evaluation tasks. We used embedding models trained by a knowledge embedding approach which has been evaluated with clinical datasets. Experimental results with Pearson and Spearman correlations show strong evidence that the novel proposed evaluation protocol is pottentially able to replace link prediction. 
\end{abstract}



\section{Introduction}



In the last decade, with the development of knowledge graphs (KGs), the attention to multi-relational data has been growing rapidly. KGs, such as Google Knowledge Vault~\cite{Dong:2014} and Freebase~\cite{Bollacker:2008}, are  multi-relational graphs composed by entities (nodes) and relations (edges) that connect heterogeneous information together and provide the ability to analyse problems from a relationship perspective in machine learning related applications, such as information retrieval 
~\cite{nikolaev2016parameterized}, semantic parsing~\cite{Berant:2013}, recommender systems~\cite{Wang:2019} and question answering~\cite{Abujabal:2017}.
%
KG constituents include a triplet subject-predicate-object, usually denoted as head, relation and tail. Knowledge embedding models are used to translate the relations among entities from the head entity to tail entity, favouring to keep the intrinsic semantics of the constituents. 

In the clinical domain, the proliferation of electronic health records presents both opportunities and challenges for accelerating the large-scale manipulation of clinical data by data science. The result is a wealth of information that combines structured and unstructured clinical data, 
inspiring the need to develop appropriate computing techniques for its analysis. The increasing number of multi-relational data also means that the evaluations of embedding representation models need to be carefully considered as long as minimal errors can lead to enormous negative impact in AI related applications. 
Hence, a higher requirement for model evaluation is necessary as any tiny residuals or loss of data will be amplified in a wide range of applications.

Given the growth of structured data in health related  systems, it has become harder for domain experts to manually analyse and utilise the information contained in these systems in an efficient and timely manner.
There are a variety number of clinical data of which the main components (e.g. diagnosis, prescription and physician) of the data form a typical multi-relational data structure. Learning the knowledge low-dimensional embedding representation for entities and relations extracted from these types of datasets is meaningful. 
Potentially important patterns can be hidden in the large number of relations returned by querying these data sources. To alleviate these problems,  
embedding KG constituents is a potential solution for detecting patterns in order to improve healthcare processes. In addition, KG can be applied to the content of electronic medical record databases to uncover meaningful associations between the most diverse sources of clinical data.


TransE~\cite{2013_Bordes_TransE} is the baseline representative embedding model that learns vector embeddings for both entities and relations. For a triple $(h,r,t)$ which represents the relation $r$ between a head entity $h$ and a tail entity $t$, the basic idea of TransE model is that this triple induce a functional relation corresponding to a translation of the embedding of entities in a vector space.
TransE solves the problem of too many parameters in the previous models, but it is not designed to model and correctly deal with one-to-many and many-to-many relations. Following TransE, a series of works are proposed to address those shortcomings, such as 
TransH~\cite{2014_Wang_TransH}, TransR~\cite{2015_Lin_TransR},  TransD~\cite{ji2015knowledge}, among many others.
Link prediction (LP) is the usual TransE-like validation protocol that aims to predict missing head or tail entities in $(h,r,t)$ triples, usually using three validation metrics: Mean Reciprocal Rank (MRR), Mean Rank (MRank) and Hits@N~\cite{Xie2016}.
The commonality of these three metrics is that the evaluation of the embedding process relies on the benchmark dataset to be split into training, and multiple evaluation  sets.
However, in the clinical domain, KGs are not necessarily incomplete, and using the entire dataset during the training process could be a way of improving the latent representation and avoid loss of information.



The amount of data used for verification and testing can typically account for approximately 20\% of all the data~\cite{guyon1997scaling} and this proportion of data could provide additional effective contribution for the model training process, if not kept aside for validation purposes. 
%
Thus, it is meaningful to reduce the loss of information caused by dataset splitting if the accuracy of the model and the validity and generalisation of the evaluation can be guaranteed by an alternative validation protocol.




In this work, we propose a \emph{Clustering Evaluation Protocol} (CEP) based on the K-means algorithm that aims to replace LP as the traditional evaluation protocol along the knowledge embedding representation process for domain-specific multi-relational data. We compare the performance between clustering and link prediction evaluation protocols, and we demonstrate the feasibility of such replacement by analysing the correlation between accuracy metrics from both evaluation protocols.  

\section{Background}

\subsection{Embedding Representation}

TransE~\cite{Bordes:2013} is a baseline model that uses simple assumptions  along the  knowledge  embedding  process.
TransE learns vectors for both entities and relations, so that the relationship between two entities corresponds to a translation between the embeddings of such entities that holds the similarity described by Equation~\ref{equation:TransE}.

\begin{equation}\label{equation:TransE}
h + r \approx t
\end{equation}

Although TransE is relatively efficient in achieving predictive performance, it has some limitations when dealing with certain types of relationships, including one-to-many and many-to-many relations~\cite{2014_Wang_TransH}. 
Thus, other methods utilise TransE as an initialisation model, then enhancing TransE model with more complex embedding components in order to better capture semantics and relationships among entities, e.g. by pre-projecting entities into the relational space using a relation-specific matrix, including TransH~\cite{2014_Wang_TransH}, TransR~\cite{2015_Lin_TransR}, and TransD~\cite{ji2015knowledge}.

TransH~\cite{2014_Wang_TransH} transforms the relation into a hyper-planes and trying to translate on the hyper-planes. By keeping the mapping properties of relations, and keeping the same running time and model complexity of TransE, TransH overcomes the drawbacks of TransE regarding relation types. When different relationships are involved, each entity can have a different distributed representation, which allows entities to play different roles in different relationships. Each relation \emph{r} is represented by a vector $r$ on a hyper-plane with $w_r$ as the normal vector. The entity embedding vectors \emph{h} and \emph{t} are first projected to the hyper-plane of $w_r$ ($h_{\perp}$ and $t_{\perp}$). The score function is similar to that used in TransE, but using the projected embedding vectors instead (Equation~\ref{equation:TransH}).

\begin{equation}\label{equation:TransH}
f_r(h,t) = \lVert h_{\perp} + r - t_{\perp} \rVert_2^2
\end{equation}

TransR~\cite{2015_Lin_TransR} and ETransR~\cite{2017_Lin_ETransR} model entities and relationships are embedded into separate different vector spaces, bridged by a relation-specific matrix $M_r$ (a $k$-dimensional space of the entity and the $m$-dimensional space of the relationship). These methods focus primarily on modelling individual knowledge in a contiguous space rather than the semantic relevance between modelling knowledge. In these models, the entity and relationship embedding dimensions are not necessarily to be same. However, in ETransR, all results report $k =m$, which leads us to the following conclusions: (a) Entity embedding entities into low-dimensional spaces may lose some valuable information, and (b) use higher dimensional space does not necessarily add any other useful information to the embedded model.




TransD~\cite{ji2015knowledge} considers the different types of entities and relations synchronously. Each relation-entity pair $(r, e)$ will have a mapping matrix Mre to map entity embedding into relation vector space. And the projected vectors could be defined as $h_{\perp} = M_{rh}h$ and $t_{\perp}= M_{rt}t$. The loss function of TransD is the same used in TransH.


Alternatively, an entity-type centred perspective is presented by TransT~\cite{2017_TransT}, which tries to describe the categories of entities by combining entity types and structured information. TransT builds relation types according to entity types and relative semantic similarity, which capture prior distributions of entities and relations. But, when it produces embedding representations of each entity from different contexts, the embedding representations are different.

Type-based constraints can support the statistical modelling with latent variable models, by integrating prior knowledge on entity and relation types, significantly improving these models up to about 70\% in link prediction tasks~\cite{krompassISWC2015}, especially when a low model complexity is enforced. 
HEXTRATO~\cite{2018_Hegler_HEXTRATO} 
couples the simple assumptions from TransE with typed-based entities inspired by TransT, and a set of ontology-based constraints to learn representations for domain-specific categorical multi-relational data, having been originally designed to operate with datasets in the clinical domain. 

In categorical datasets, each entity $e$ is associated with a category (or type) $c \in \mathcal{T}$. HEXTRATO creates a set of independent type-based hyperspaces in order to project each entity belonging to the same type, allowing more space to be used along the embedding process, while speeding up the validation process by restricting the set of entities to be used on creating corrupted triples. 

HEXTRATO improves the translational embedding from TransE and other enhanced models, achieving great performance on the LP task, even in very low $k$-dimensional spaces ($k < 50$), without necessarily adding complex representation structures within the model training process, such as those used in TransE-based enhanced models.

\subsection{Link Prediction}


The traditional link prediction 
task~\cite{2013_Bordes_TransE} is used to estimate the translating embedding method accuracy. LP is focused on the prediction of a missing head or tail entity in $(h, r, t)$, from which accuracy is given by Mean Rank (MRank) and Mean Reciprocal Rank (MRR) -- a percentage of correct entities ranked in the top N can also be estimated (Hits@N, with related work usually reporting $N=10$). A good link predictor should achieve lower MRank or higher MRR. 

Along LP, we replace the original head or tail entities in each triple $(h, r, t)$ within the evaluation set, replacing $h$ or $t$ with all the entities in the dictionary in order to produce corrupted triples. Then we calculate the score function $f_r(h,t)$ in order to get all the potential tail that we could expect from the all the fake triples. In this way, we can get a series of scores that can be sorted in ascending order. The average rank of the correct original triple gives the MRank accuracy. MRR is the average of inverse rank. However, one of the drawbacks of LP is that each triple has to be compared to corrupted triples produced with all the entities from the KG, which harms the embedding processing regarding time processing.

\subsection{K-means}
K-means~\cite{macqueen1967some} is an unsupervised clustering method used to partition a dataset into K groups. This method does not need training labelled data and the process takes the following steps:
(1) the number of clusters $K$ is defined based on research needs and selecting $K$ initial cluster centres is a pre-processing step;
(2) the distance between each entity $d$ and all the initial cluster centres is calculated, and instances are assigned to their closest cluster centre;
(3) the initial cluster centres are replaced by the mean of its constituent entities; finally, 
(4) the new distances between each entity $d$ and the new cluster centres are calculated, and each entity $d$ is reassigned to the closest cluster centre.

The algorithm converges when the clustering results no longer changes after repeating steps 3 and 4. 
K-means is originally used for unlabelled data and helps finding groups of potential similar entities within a dataset.

\section{Method}

When learning embedding representation for KGs, the LP task is traditionally used as the evaluation protocol.
Such evaluation process is a time-consuming task that makes the training process longer, thus allowing each validation cycle occurs usually only after each 20-50 epochs. Most importantly, it requires splitting the original set of triples from the KG into training,  validation and test sets, which leads to loss of information. 
In order to overcome such obstacles, we evaluated the potential of K-means as an alternative validation protocol alongside the training process in embedding representation models.

We assume that entities take place in the hyperspace tending to be somehow aggregated by some of the inherited semantic similarity captured from the original multi-relational data (e.g. similar drug effects, or patients with similar clinical conditions). Although each entity can participate in multiple relations, those groups can still capture some of the original features as long as the model improves and tries to reach better embedding representation. Therefore, clustering a set of target entities based on their vector representation is expected to result group wise clusters similar to the organization of the original entity labels. 

The feasibility of unsupervised learning in the assessment process has been applied to some extent~\cite{yuan2019constructing}. Using unsupervised learning methods to evaluate the embedding process is a potential way to overcome some of the issues in the original LP task, such as information loss due to dataset splitting. 
%


\subsection{Datasets}

\begin{table}[t]
\caption{Statistics of clinical benchmark datasets, given by the number of entities, relations, types, and triples in each dataset split -- training (LRN), validation (VLD), tuning (TUN) and test (TST) sets.}
\label{tab:datasets-stats}
\center
\begin{tabular} { | p{2cm} | p{2.5cm} | p{2.5cm} | p{2.5cm} | }
\hline											
		&	\multicolumn{3}{c|}{\bf{EHR Datasets}}	\\
\cline{2-4} 
\bf{\# (number of)}		&	\multicolumn{1}{c|}{\bf{Demographics}}	&	\multicolumn{1}{c|}{\bf{Pregnancy}}   & \multicolumn{1}{c|}{\bf{BPA}}  \\
\hline											

Entities 		& \multicolumn{1}{r|}{2,237}	& \multicolumn{1}{r|}{3,088	}	&  \multicolumn{1}{r|}{22,874}  	\\
\hline											
Relations		& \multicolumn{1}{r|}{6}		& \multicolumn{1}{r|}{5}		& \multicolumn{1}{r|}{23  } 	\\
\hline											
Types			& \multicolumn{1}{r|}{7}		& \multicolumn{1}{r|}{4}		& \multicolumn{1}{r|}{14 }	\\
\hline											
Triples (total) & \multicolumn{1}{r|}{15,345}	& \multicolumn{1}{r|}{20,768}	& \multicolumn{1}{r|}{186,177}	\\
\ \ \ \ \ LRN	& \multicolumn{1}{r|}{13,875}	& \multicolumn{1}{r|}{14,588}	& \multicolumn{1}{r|}{177,727}	\\
\ \ \ \ \ VLD	& \multicolumn{1}{r|}{463}		& \multicolumn{1}{r|}{1,997	}	& \multicolumn{1}{r|}{2,889 }	\\
\ \ \ \ \ TUN	& \multicolumn{1}{r|}{475}		& \multicolumn{1}{r|}{2,093}	& \multicolumn{1}{r|}{2,729 }	\\
\ \ \ \ \ TST	& \multicolumn{1}{r|}{532}		& \multicolumn{1}{r|}{2,090}	& \multicolumn{1}{r|}{2,832} 	\\
\hline											
\end{tabular}
\end{table}

We used two real clinical datasets originally used in the HEXTRATO benchmark (Demographics and Pregnancy) and an additional clinical-related ontology dataset (BPA), all obtained from \emph{InfoSaude}~\cite{Tissot2018MND,Tissot2019ITN}, an Electronic Health Record (EHR) system used to manage patient records,
such as outpatient information and type of care, pregnancies and drug prescriptions.\footnote{links the all datasets, including de-identified versions of EHR datasets, will be made available upon acceptance.} 
Further details about each dataset are presented below and statistics are depicted in Table~\ref{tab:datasets-stats}.

{\bf EHR-Demographics}
comprises a set of 2,185 randomly selected patients from the \emph{InfoSaude} system who had at least one admission between 2014 and 2016. Each patient is described by a set of basic demographic information, including gender, age (range in years) in the admission, marital status (unknown for about 15\% of the patients), education level, and two flags indicating whether the patient is known to be either a smoker or pregnant, and the social groups assigned according to a diverse set of rules mainly based on demographic and historical clinical conditions. Demographics features are represented by \emph{many-to-one} relations, whereas association of each patient to social groups is given by a many-to-many relation.

{\bf EHR-Pregnancy}
is a dataset used to identify correlations between pre- and post-clinical conditions on pregnant patients with abnormal pregnancy termination, comprised by a set of 2,879 randomly selected pregnant female patients from the \emph{InfoSaude} system which pregnancy was inadvertently and abnormally interrupted before the expected date of birth; each patient is described by age (range in years),  known date of last menstrual period (LMP), whether the patient had an abortion (regardless of reason), and a list of \mbox{ICD-10} (the 10th revision of the International Classification of Diseases) codes~\cite{WHOICD10} registered either before or after the LMP date. This is mostly a dataset comprising \emph{many-to-many} relations that connects patients with corresponding diagnoses.

{\bf BPA} (Ambulatory Production Bulletin)\footnote{\url{http://datasus.saude.gov.br/sistemas-e-aplicativos/ambulatoriais/sia}}
is an outpatient care dataset that allows the service provider to be linked to the Public Health Ministry in Brazil to record the care performed at the health facility on an outpatient basis; in order to optimize the data remittance process, there are several rules for the correct completion of submitted data that must be followed strictly, including the restrictions between medical procedures and their constrained diagnoses from \mbox{ICD-10}. This dataset associates medical procedures with a multilevel hierarchical set of many-to-one relations, coupled with many-to-many relations that impose multiple restrictions on each procedure, mainly regarding to possible related diagnoses.

\subsection{Learning Embeddings}

We used HEXTRATO to pre-train a set of distinct models for each dataset, by combining different values of hyper parameters. 
The set of golden triples is then randomly traversed multiple times along the training process up to the maximum of 1,000 iterations, such that each training step produces a corrupted triple for each correct triple.
HEXTRATO uses a \emph{tuning} set to choose the best of multiple replicas independently initialised with random vector representations for each entity and relation. 
Final resulting LP scores are then calculated over the test set.

The combination of hyper parameters used for training multiple models include margin parameter $\gamma \in \{1.0, 2.0, 4.0\}$, the set of four cumulatively combined ontology-based constraints proposed by HEXTRATO, among others, tested in distinct $k$-dimensional spaces, where $k \in \{16,32,64\}$, resulting a total of 1,728 embedding models for each dataset. 
%
For each set of hyper parameters used to train each model, resulting MRR and MRank varies in each dataset accordingly (Table~\ref{tab:min_max_LP}). In BPA and Pregnancy, MRank is not favoured due to the predominance of many-to-many relations, and LP model accuracy is highly susceptible to the choice of hyper parameters. 

\begin{table}[t]
\begin{center}
\caption{Ranges of resulting MRank and MRR in each datasets.}
\label{tab:min_max_LP}
\begin{tabular}{|l|r|r|r|r|r|r|}
\hline
& \multicolumn{2}{c|}{\bf{k = 32}} & \multicolumn{2}{c|}{\bf{k = 64}} \\ \cline{2-5} 
\bf{Dataset}                  & \bf{MRR}             & \bf{MRank}            & \bf{MRR}             & \bf{MRank}            \\ \hline
Demographics       & 0.592           & 3.080            & 0.597           & 3.097            \\ \hline
Pregnancy         & 0.346           & 24.226           & 0.351           & 23.236           \\ \hline
BPA               & 0.771           & 35.265           & 0.781           & 28.969           \\ \hline
\end{tabular}
\end{center}
\end{table}

Each combination of hyper parameters is more or less susceptible to produce noise along the embedding training process, either favouring or harming the quality of resulting vector representation. This give us a wider range of accuracy values that made it possible to compare the resulting embeddings 
against the proposed evaluation protocol based on clustering.

\subsection{Clustering Evaluation Protocol}

For each dataset, we targetted one specific type of entities, and used multiples entity feature labels to  evaluate clustering. Along initial experiments with multiple numbers of clusters, we tested $L$, $2L$, and $4L$ target clusters separately, where $L$ represents the number of the distinct values in each target label set. We found that increasing the numbers of target clusters favours the accuracy of resulting groups, similarly mimicking the original way in which feature labels separate the set of entities. Thus, accuracy for $4L$ clusters is reported in Section~\ref{section:Results}.



We compare whether the accuracy of the clustering method based on each target relation is correlated to the accuracy given by MRank and MRR 
metrics from the LP task. 
For each dataset, we perform cluster analysis separately on the resulting embedding models, so the number of clustering is pairwise with number of models.
%
The target entity type used for clustering and the set of target labels obtained from relation in each KG are described in Table~\ref{tab:clusteringgroup}. 

\begin{table}
\begin{center}
\caption{Number of classes in each target clustering relation.}
\label{tab:clusteringgroup}
\begin{tabular}{|l|l|l|c|}
\hline
     & \bf{Target} & \bf{Target} &  \\ 
\bf{Dataset}     & \bf{Type} & \bf{Relation} & \bf{\#Classes}  \\ \hline
Demographics & Patient & hasMaritalStatus  & 4       \\
            &         & ageStage  & 6       \\ \hline
Pregnancy   & Patient & hadAbortion  & 2       \\ 
            &         & ageWeeksPregnancyInterrupted  & 15       \\ \hline
BPA         & Procedure & PGroup    & 8       \\ 
            &           & PSubgroup & 59      \\ \hline
\end{tabular}
\end{center}
\end{table}


In order to evaluate in what extent CEP can be used as a replacement metric to the usual LP task, we compare the clustering result from each target label against the result embedding model LP accuracy given by MRank and MRR. 
To calculate the K-means accuracy, we use the predominant target label in each cluster as a intra-cluster accuracy ($Acc_k$). Then we calculate the arithmetic and weighted means as taken by accuracy metrics (Equations~\ref{equation:Arithmetic} and~\ref{equation:Weighted}).





\begin{equation}\label{equation:Arithmetic}
aMean=\frac{\sum_{k=1}^{N}Acc_{k}}{N}
\end{equation}

\begin{equation}\label{equation:Weighted}
wMean=\frac{\sum_{k=1}^{N}Acc_{k}*\frac{T_{k}}{L}}{\sum_{k=1}^{N}\frac{T_{k}}{L}}
\end{equation}

The arithmetic mean does not take the distribution of original labels into account, whereas the weighted mean considers the number of entities in order to weight each cluster. In the later, the degree of influence in which each cluster accuracy contributes to the overall accuracy rate depends on the proportion of the target predominant labels $T_k$ in each cluster $k$ regarding the total number of target labels $L$.




\begin{figure*}[t]
  \centering
  \includegraphics[width=\textwidth]{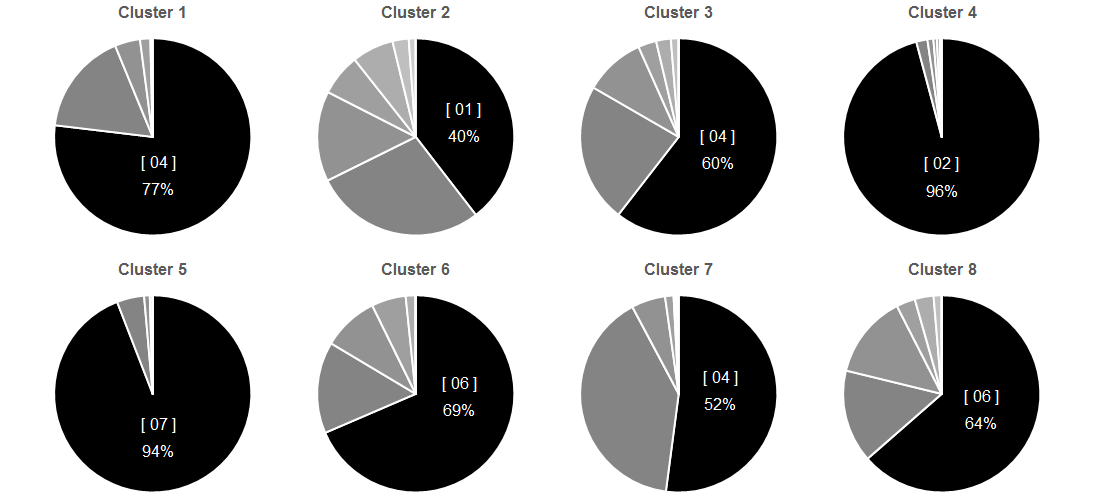} 
  \caption{Example of CEP for the BPA dataset using 8 clusters and 8 target labels -- predominant labels for each cluster are indicated as ``[$X$]" with the corresponding percentage.}
  \label{ClusterSample}
\end{figure*}

%
The resulting predominant labels from K-means does not completely coincide with the set of target labels because, regardless of the target type of each entity, unbalanced distributions of labels are not completely captured along clustering. Moreover, we use $4L$ target clusters to evaluate $L$ target labels, so that it is expected multiple clusters having the same predominant target label, whereas less representative labels in the KG being kept out of the accuracy means.

Fig.~\ref{ClusterSample} shows an example of a clustering setup for the BPA dataset targeting the PGroup label for each medical procedure (we used $L=8$ clusters instead of $4L$ just for simplification). 
The target entity labels range from 01 to 08, and resulting clusters are numbered from 1 to 8. In this example, 
each target label is somehow distributed in each cluster, but one label is always predominant. Labels 04 and 06 are repeatedly predominant in multiple clusters, whereas labels 03 and 05 are not representative for accuracy purposes. 


%
%


\section{Results}
\label{section:Results}

For each dataset, we used HEXTRATO to build a set of 1,728 models, each one corresponding to a distinct combination of initialisation hyper parameters. Each model outputs the embedding vectorial representation for each entity and relation that compound each dataset.
%
Using the resulting embeddings and labels for each entity as the input for the clustering model, we obtained the clustering result accuracy of each embedding model. Then, we compare LP and CEP results, by correlating MRank and MRR from LP against CEP accuracy metrics.

To contrast CEP and LP, we compared Pearson and Spearman correlations between each evaluation metric from both methods. Pearson's tends to assess linear relationships, whereas Spearman's tends to assess monotonic relationships. The Spearman's coefficient is also defined as the Pearson's correlation coefficient between the rank variables.
Pearson and Spearman correlations between LP and CEP are shown in Table~\ref{tab:CorrelationResults}.
By comparing the LP metrics MRR and MRank against CEP metrics aMean and wMean, we found both correlations are consistent to each other within each dataset, but they differ according to the shape of each one.
%






BPA dataset comprises a 4-level hierarchical  structure of medical procedures given by a set of nested many-to-one relations (\emph{inPGroup}, \emph{inPSubgroup}, \emph{inPOrgForm}), balanced with a many-to-many relation that defines multiple restrictions to each medical procedure (\emph{isRestrictedBy}) with multiple range types. The strong resulting correlation is in line with our expectations as the effect of CEP increases with the increase of MRank and MRR.
%



Unexpectedly, for Pregnancy and Demographics datasets, LP and CEP metrics are mostly uncorrelated to each other. This can be explained due to two possible factors: 
(a) in both Demographics and Pregnancy datasets we observed  high accuracy repeatedly resulting from CEP, which is not consistent with relatively low accuracy given the LP metrics; alternatively (b) both datasets are predominantly compound by many-to-many relations, whereas we used one-to-many relations to perform the proposed clustering protocol, and the predominant relations can be overlaying the semantic relationship contribution given by the other relations.



\begin{table*}[t]
\begin{center}
\caption{Resulting Pearson and Spearman correlation between LP and CEP metrics.}
\label{tab:CorrelationResults}
\begin{tabular}{|l|l|l|r|r|r|r|}
\hline
& \bf{Target} &	\bf{LP} &	\multicolumn{2}{c|}{\bf{CEP Pearson's}}&	\multicolumn{2}{c|}{\bf{CEP Spearman's}} \\ \cline{4-7}

\bf{Dataset}     & \bf{Label}		& \bf{Metric}	&	aMean           & wMean           & aMean                      & wMean            \\ \hline
BPA         & PGroup 	& MRR	& 0.776 & 0.773 & 0.715 & 0.709 \\ \cline{3-7}
			& 		 	& MRank	& -0.635 & -0.631 & -0.669 & -0.656\\ \cline{2-7}
			& PSubgroup & MRR 	& 0.769 & 0.780 & 0.723 & 0.732 \\ \cline{3-7}
			& 		 	& MRank & -0.572 & -0.584 & -0.648 & -0.661\\ \hline
Pregnancy   & hadAbortion 	& MRR	& 	-0.178 & -0.188 & -0.172 & -0.178\\ \cline{3-7}
			& 		 	& MRank	& 		0.124 & 0.125 & 0.126 & 0.123 \\ \cline{2-7}
			& ageWeeksPregnancyInterrupted & MRR 	& 0.091 & 0.005 & 0.038 & -0.032 \\ \cline{3-7}
			& 		 	& MRank & 		-0.211 & -0.123 & -0.160 & -0.080 \\ \hline
Demographics& hasMaritalStatus 	& MRR& 	-0.074 & -0.088& -0.065 & -0.073 \\ \cline{3-7}
			& 		 	& MRank	& 		0.031 & 0.045 & 0.010 & 0.017 \\ \cline{2-7}
			& ageStage & MRR 	& 		0.211 & 0.195 & 0.062 & 0.056 \\ \cline{3-7}
			& 		 	& MRank & 		-0.133 & -0.125 & -0.092 & -0.086\\ \hline
\end{tabular}
\end{center}
\end{table*}

Correlations between MRank and CEP are expected to be negative, whereas correlations between MRR and CEP are expected to be negative.
However, we found contradictory results in both EHR  datasets that led to further reflection about whether the shape of each dataset suits better for distinct evalution protocols. 
In addition, we found a tendency to the correlation becomes impaired as the smaller is the size of datasets.  
\section{Conclusions}

We propose a Cluster Evaluation Protocol (CEP) that aims to replace the traditional link prediction (LP) as the evaluation protocol of knowledge embedding representation for categorised multi-relational data in the clinical domain. By using the knowledge embedding approach HEXTRATO on two clinical-related datasets and one clinical ontology over different hyper parameter combinations, vector representation for each entity was used as input in a clustering algorithm. The accuracy of clustering based on multiple target labels was contrasted to LP metrics MRR and MRank, by using Pearson and Spearman correlation coefficients. 


Experimental benchmark results show that the correlation between CEP and LP varies according to the size and shape (predominant relation types). In an ontology-like dataset with a certain balance between one-to-many and many-to-many relations, the correlation between LP and CEP scores is high, which supports our primary aim of using CEP as an embedding evaluation protocol.

In contrast, we observed the same correlation is weak for EHR datasets in which many-to-many relations are most predominantly compounding the datasets. Thus, further investigation is required in order to establish whether the observed weak correlation results from (a) the high accuracy of CEP, or (b) the dataset shapes.
Alternatively, there is a risen question about the quality of the resulting embedding representation. The original LP evaluation focuses on completing knowledge graphs whereas CEP is designed to obtain the best vectorial representation as possible in order to embed the semantic relationships between entities. 

Some of the directions in which this work can be extended include:
%
(a) further analysis in order to compare the effectiveness of specific initial hyper parameters in the CEP accuracy; 
(b) contrast other ontology-based constraints provided by HEXTRATO and how they can influence either LP or CEP as embedding evaluation protocols for clinical-related datasets;
(c) further experimental exercises in order to establish a connection between LP and/or CEP with the quality of the resulting embedding representation, possibly by correlating the embedding evaluation protocol metrics with other metrics resulting from classification tasks that use  entity embeddings as input.



\bibliographystyle{ieeetr}
\bibliography{main.bib}

\begin{thebibliography}{10}

\bibitem{Dong:2014}
X.~Dong, E.~Gabrilovich, G.~Heitz, W.~Horn, N.~Lao, K.~Murphy, T.~Strohmann,
  S.~Sun, and W.~Zhang, ``Knowledge vault: A web-scale approach to
  probabilistic knowledge fusion,'' in {\em Proceedings of the 20th ACM SIGKDD
  international conference on Knowledge discovery and data mining},
  pp.~601--610, ACM, 2014.

\bibitem{Bollacker:2008}
K.~Bollacker, C.~Evans, P.~Paritosh, T.~Sturge, and J.~Taylor, ``Freebase: A
  collaboratively created graph database for structuring human knowledge,'' in
  {\em Proceedings of the 2008 ACM SIGMOD International Conference on
  Management of Data}, SIGMOD '08, (New York, NY, USA), pp.~1247--1250, ACM,
  2008.

\bibitem{nikolaev2016parameterized}
F.~Nikolaev, A.~Kotov, and N.~Zhiltsov, ``Parameterized fielded term dependence
  models for ad-hoc entity retrieval from knowledge graph,'' in {\em
  Proceedings of the 39th International ACM SIGIR conference on Research and
  Development in Information Retrieval}, pp.~435--444, ACM, 2016.

\bibitem{Berant:2013}
J.~Berant, A.~Chou, R.~Frostig, and P.~Liang, ``Semantic parsing on freebase
  from question-answer pairs,'' in {\em Proceedings of the 2013 Conference on
  Empirical Methods in Natural Language Processing}, pp.~1533--1544, 2013.

\bibitem{Wang:2019}
H.~Wang, F.~Zhang, J.~Wang, M.~Zhao, W.~Li, X.~Xie, and M.~Guo, ``Exploring
  high-order user preference on the knowledge graph for recommender systems,''
  {\em ACM Trans. Inf. Syst.}, vol.~37, pp.~32:1--32:26, Mar. 2019.

\bibitem{Abujabal:2017}
A.~Abujabal, M.~Yahya, M.~Riedewald, and G.~Weikum, ``Automated template
  generation for question answering over knowledge graphs,'' in {\em
  Proceedings of the 26th international conference on world wide web},
  pp.~1191--1200, International World Wide Web Conferences Steering Committee,
  2017.

\bibitem{2013_Bordes_TransE}
A.~Bordes, N.~Usunier, A.~Garcia-Duran, J.~Weston, and O.~Yakhnenko,
  ``Translating embeddings for modeling multi-relational data,'' in {\em
  Advances in Neural Information Processing Systems 26} (C.~J.~C. Burges,
  L.~Bottou, M.~Welling, Z.~Ghahramani, and K.~Q. Weinberger, eds.),
  pp.~2787--2795, Curran Associates, Inc., 2013.

\bibitem{2014_Wang_TransH}
Z.~Wang, J.~Zhang, J.~Feng, and Z.~Chen, ``Knowledge graph embedding by
  translating on hyperplanes.,'' in {\em Proceedings of the Twenty-Eighth
  {AAAI} Conference on Artificial Intelligence} (C.~E. Brodley and P.~Stone,
  eds.), pp.~1112--1119, {AAAI} Press, 2014.

\bibitem{2015_Lin_TransR}
Y.~Lin, Z.~Liu, M.~Sun, Y.~Liu, and X.~Zhu, ``Learning entity and relation
  embeddings for knowledge graph completion,'' in {\em Proceedings of the
  Twenty-Ninth AAAI Conference on Artificial Intelligence}, AAAI'15,
  pp.~2181--2187, AAAI Press, 2015.

\bibitem{ji2015knowledge}
G.~Ji, S.~He, L.~Xu, K.~Liu, and J.~Zhao, ``Knowledge graph embedding via
  dynamic mapping matrix,'' in {\em Proceedings of the 53rd Annual Meeting of
  the Association for Computational Linguistics and the 7th International Joint
  Conference on Natural Language Processing (Volume 1: Long Papers)}, vol.~1,
  pp.~687--696, 2015.

\bibitem{Xie2016}
R.~Xie, Z.~Liu, J.~Jia, H.~Luan, and M.~Sun, ``Representation learning of
  knowledge graphs with entity descriptions,'' in {\em Proceedings of the
  Thirtieth AAAI Conference on Artificial Intelligence}, AAAI'16,
  pp.~2659--2665, AAAI Press, 2016.

\bibitem{guyon1997scaling}
I.~Guyon, ``A scaling law for the validation-set training-set size ratio,''
  {\em AT\&T Bell Laboratories}, pp.~1--11, 1997.

\bibitem{Bordes:2013}
A.~Bordes, N.~Usunier, A.~Garcia-Duran, J.~Weston, and O.~Yakhnenko,
  ``Translating embeddings for modeling multi-relational data,'' in {\em
  Advances in Neural Information Processing Systems 26} (C.~J.~C. Burges,
  L.~Bottou, M.~Welling, Z.~Ghahramani, and K.~Q. Weinberger, eds.),
  pp.~2787--2795, Curran Associates, Inc., 2013.

\bibitem{2017_Lin_ETransR}
H.~Lin, Y.~Liu, W.~Wang, Y.~Yue, and Z.~Lin, ``Learning entity and relation
  embeddings for knowledge resolution,'' {\em Procedia Computer Science},
  vol.~108, no.~Supplement C, pp.~345 -- 354, 2017.
\newblock International Conference on Computational Science, ICCS 2017, 12-14
  June 2017, Zurich, Switzerland.

\bibitem{2017_TransT}
S.~Ma, J.~Ding, W.~Jia, K.~Wang, and M.~Guo, ``Trans{T}: Type-based multiple
  embedding representations for knowledge graph completion,'' in {\em The
  European Conference on Machine Learning and Principles and Practice of
  Knowledge Discovery in Databases}, 2017.

\bibitem{krompassISWC2015}
D.~Krompa{\ss}, S.~Baier, and V.~Tresp, ``Type-constrained representation
  learning in knowledge graphs,'' in {\em Proceedings of the 13th International
  Semantic Web Conference (ISWC)}, 2015.

\bibitem{2018_Hegler_HEXTRATO}
H.~Tissot, ``{HEXTRATO:} using ontology-based constraints to improve accuracy
  on learning domain-specific entity and relationship embedding representation
  for knowledge resolution,'' in {\em {IC3K 2018 10th International Joint
  Conference on Knowledge Discovery, Knowledge Engineering and Knowledge
  Management}}, vol.~1, pp.~72--81, SciTePress, 2018.

\bibitem{macqueen1967some}
J.~MacQueen {\em et~al.}, ``Some methods for classification and analysis of
  multivariate observations,'' in {\em Proceedings of the fifth Berkeley
  symposium on mathematical statistics and probability}, vol.~1, 14,
  pp.~281--297, Oakland, CA, USA, 1967.

\bibitem{yuan2019constructing}
J.~Yuan, Z.~Jin, H.~Guo, H.~Jin, X.~Zhang, T.~Smith, and J.~Luo, ``Constructing
  biomedical domain-specific knowledge graph with minimum supervision,'' {\em
  Knowledge and Information Systems}, pp.~1--20, 2019.

\bibitem{Tissot2018MND}
H.~Tissot and R.~Dobson, ``Identifying misspelt names of drugs in medical
  records written in portuguese,'' {\em {HealTAC}-2018: Unlocking Evidence
  Contained in Healthcare Free-text}, 2018.

\bibitem{Tissot2019ITN}
H.~Tissot, M.~D. Del~Fabro, L.~Derczynski, and A.~Roberts, ``Normalisation of
  imprecise temporal expressions extracted from text,'' {\em Knowledge and
  Information Systems}, Feb 2019.

\bibitem{WHOICD10}
W.~H.~O. WHO, {\em ICD-10 : international statistical classification of
  diseases and related health problems / World Health Organization}.
\newblock World Health Organization Geneva, 10th revision, 2nd ed.~ed., 2004.

\end{thebibliography}

\end{document}